\documentclass{article}

     \usepackage[nonatbib, final]{neurips_2019}
\usepackage[numbers]{natbib}

\usepackage[utf8]{inputenc} 
\usepackage[T1]{fontenc}    
\usepackage{hyperref}       
\usepackage{url}            
\usepackage{booktabs}       
\usepackage{amsfonts}       
\usepackage{nicefrac}       
\usepackage{microtype}      
\usepackage{subfig}
\usepackage{graphicx}
\usepackage{xcolor}

\title{Machine Learning for Generalizable Prediction\\of Flood Susceptibility}

\author{%
  Chelsea Sidrane\thanks{These authors contributed equally to this work.}\\
  Stanford University\\
  \texttt{csidrane@stanford.edu } \\
  \And
  Dylan J Fitzpatrick\textsuperscript{*}\\
  Carnegie Mellon University\\
  \texttt{djfitzpa@cmu.edu} \\
  \And
  Andrew Annex\textsuperscript{*}\\
  Johns Hopkins University\\
  \texttt{aannex1@jhu.edu} \\
  \And
  Diane O'Donoghue\\
  kx\\
  \texttt{dodonoghue@kx.com} \\
  \And
  Yarin Gal\\
  University of Oxford\\
  \texttt{yarin.gal@cs.ox.ac.uk} \\
  \And
  Piotr Bili\'nksi\\
  University of Warsaw\\
  \texttt{bilinski@mimuw.edu.pl} \\
}

\begin{document}

\maketitle

\begin{abstract}
Flooding is a destructive and dangerous hazard and climate change appears to be increasing the frequency of catastrophic flooding events around the world. Physics-based flood models are costly to calibrate and are rarely generalizable across different river basins, as model outputs are sensitive to site-specific parameters and human-regulated infrastructure. In contrast, statistical models implicitly account for such factors through the data on which they are trained. Such models trained primarily from remotely-sensed Earth observation data could reduce the need for extensive in-situ measurements. In this work, we develop generalizable, multi-basin models of river flooding susceptibility using geographically-distributed data from the USGS stream gauge network. Machine learning models are trained in a supervised framework to predict two measures of flood susceptibility from a mix of river basin attributes, impervious surface cover information derived from satellite imagery, and historical records of rainfall and stream height. We report prediction performance of multiple models using precision-recall curves, and compare with performance of naive baselines. This work on multi-basin flood prediction represents a step in the direction of making flood prediction accessible to all at-risk communities. 
\end{abstract}

\section{Introduction}
\label{introduction}
Among natural disasters, flooding is one of the most destructive, dangerous, and common hazards. In the U.S., 75\% of all Presidential disaster declarations are associated with flooding, and floods cause an average of \$6 billion in property damage per year~\cite{presidentialDisaster,USGSfloodThreat}. Additionally, climate change appears to be exacerbating the incidence of catastrophic flooding events~\cite{emdat}.
If we can improve the access to flood prediction systems, lives could be saved and damage lessened. 
In this work, we apply machine learning to the problem of forecasting river flooding hazards.

The National Oceanic and Atmospheric Administration (NOAA) predicts river levels at specific river locations up to several days in advance using hydrologic and hydraulic modeling~\cite{noaaForecasts, NWM}. These forecasts can help cities prepare for and rapidly respond to river flooding events. NOAA also prepares seasonal water forecasts at the regional spatial resolution~\cite{spring2019}. The work presented here is complementary to these efforts and explores the efficacy of statistical -- instead of hydrological -- models for the prediction and forecasting of river flooding. 

The physics-based models that NOAA uses operationally to predict river levels and issue official flood warnings have three key shortcomings which make their use time-consuming and costly, and therefore limits the number of communities they are available to. The first is that human-created structures such as dams and water diversions must be individually modeled by a hydrologist, often requiring extensive field observations under a variety of hydrological conditions. 
In contrast, statistical models can implicitly capture information about dams and water diversions through the data on which they are trained without the need to explicitly model their influence on water flow. Secondly, physics-based models must be calibrated to small geographic areas and consequently their outputs are not generalizable across different river basins. If many small models must be built, fewer sites may be modeled due to time and resource constraints, and fewer people are ultimately served. We believe that statistical models need not be limited to small geographic areas in the same way, which is why we perform multi-basin flood prediction. Finally, current physics-based predictions from NOAA rely on stream measurements taken using the USGS network of roughly 11,000 stream gauges, which each cost between \$7,000 and \$15,000 per year to maintain~\cite{norris2008qualitative}. If flood prediction can be done without reliance on these expensive in-situ measurements it could reduce costs for existing flood prediction systems as well as expand flood prediction efforts to areas that could not previously afford river gauging infrastructure. In sum, finding a means for fast and generalizable prediction of flood susceptibility has the potential for substantive positive impact in areas that lack extensive resources for disaster planning and response. 


Existing work using machine learning methods to predict flood susceptibility has also been constrained to small geographic areas. \citet{khosravi2018comparative} predict flood susceptibility in a single watershed, and ~\citet{tayfur2018flood} predict a hydrograph for a single stretch of river. \citet{assem2017urban} predict river levels at 3 stations in a single catchment. Work by \citet{kratzert2018rainfall} on estimating runoff as a function of rainfall using LSTMs suggests that machine learning models are able to generalize across many catchments in a manner that traditional models cannot, providing promising evidence that statistical flood susceptibility models need not be limited to small geographic areas. 
The work presented here explores prediction performance of generalizable, multi-basin models on two measures of flood susceptibility using data from six U.S. states across more than six river basins: South Dakota, Nebraska, South Carolina, Virginia, New York, and New Jersey. The models trained in this work rely on USGS stream gauge data for ground truth and as inputs for a subset of our experiments, but also incorporates remotely-sensed data and is a step in the direction of a flood prediction model that is less dependent on in-situ measurements. 

\section{Data}
To predict flood susceptibility, we rely on information sourced from a mix of satellite-derived and in-situ data. The USGS stream gauge network provides 
river height measurements at 15 minute intervals \cite{USGS:2019}. Data collected between 2009 and 2019 was used. Flood thresholds at stream gauge locations are determined by NOAA for four separate flood categories~\cite{NWS}. The `minor flood' threshold was used to binarize the USGS stream gauge readings for experiments that involve predicting flood occurrence. This resulted in a dataset where $5.5\%$ of the data points indicated that flooding occurred. 
Records of time to peak river levels after precipitation events were obtained from the Flooded Locations \& Simulated Hydrographs (FLASH) Project \cite{FLASH}. The time-to-peak data was split into 4 bins with roughly equal frequency, a time to peak of less than $3.12$ hours, between $3.12$ and $7.44$ hours, between $7.44$ and $18$ hours, and greater than $18$ hours, and assigned categorical labels indicating bin number. For historical rainfall data, daily total precipitation from the PRISM climate data set is used~\cite{PRISM:2019}. At each gauge location, river basin attributes affecting regional surface runoff and groundwater drainage are obtained from the EPA StreamCat dataset~\cite{StreamCat:2016}. Average upstream impervious surface cover at gauge locations is calculated from the satellite-derived National Land Cover Dataset (NLCD)~\cite{NLCD:2016}. We include elevation at each river gauge location by overlaying Shuttle Radar Topography Mission (SRTM) 30-meter resolution data \cite{USGS:2014}, and include a characteristic length parameter describing the scaling relationship between channel slope and drainage area \cite{Giachetta:2018}. All input data including rainfall, basin characteristics, elevation information, and impervious surface cover is normalized by feature to fall approximately in the range $[0,1]$ using the training set statistics. Each stream gauge had up to 10 years of data at 15 minute intervals which was summarized into monthly statistics. The entire record of each such location was randomly assigned to either the training set, validation set, or the test set to achieve a 60-20-20 training, validation, and test split.  

\section{Methodology}
\begin{figure}%
    \centering
    \includegraphics[width=.8\textwidth]{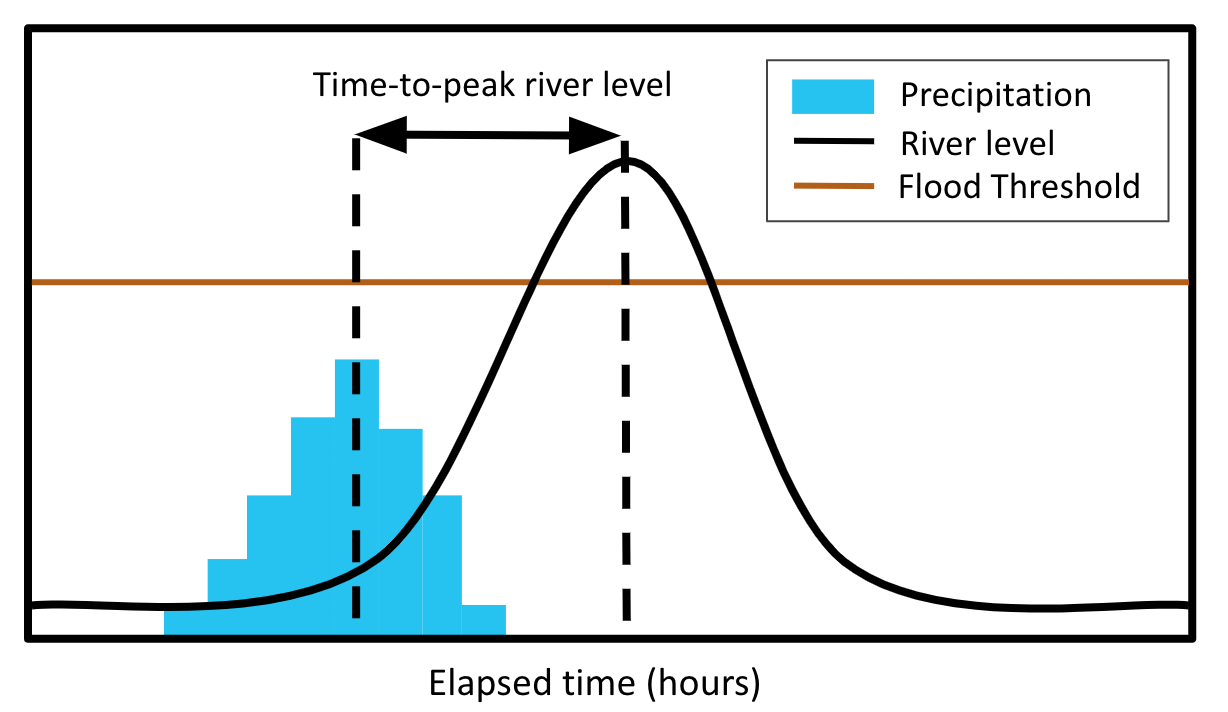}
    \caption{River height for a single stream gauge location in the period during and following a precipitation event.}%
    \label{fig:hydrograph}%
\end{figure}

We train statistical models to predict two separate measures of flood susceptibility: (1) binary flood occurrence at gauge locations within a given month (approx. 50,000 gauge-months), and (2) time to peak river level after precipitation events (approx. 3000 events). These two measures of flood susceptibility are indicated on a plot of river height over time for a single location in Figure \ref{fig:hydrograph}, where flood occurrence indicator is triggered once the river height crosses above a fixed flood threshold. Taken together, these two prediction targets provide critical information for both long-term and regional-scale flood planning, and short-term planning for localized flash flooding following extreme rainfall events. In Section \ref{sec:results}, we report results for a single time-to-peak bin (greater than 18 hours from start of precipitation to peak river level).

A key objective of this work is to explore the relative importance of different features for making accurate predictions. In particular, how critical are recent historical records of river height and accurate rainfall forecasts for predicting future flood susceptibility? These questions can help us understand the relative feasibility of statistical flood prediction for an ungauged river basin and for an area where accurate rainfall forecasts aren't available. To answer these questions, we construct three experiments that include different groups of features for the statistical models. Experiment 1 addresses the problem of predicting flood susceptibility in ungauged locations along a river network. For these experiments, no information on prior river levels is provided as input the predictive models. Experiment 2 assumes a well-gauged river basin, and forecasts flood susceptibility using historical data on river height at prediction locations. Finally, Experiment 3  provides an informative upper bound on prediction performance in the presence of accurate rainfall forecasts by including true rainfall observations 
as an input to predictive models.

Three classification models are trained in a supervised framework to predict both measures of flood susceptibility. The first, a random forest classifier, constructs an ensemble of decision trees from a bootstrap sample of the training set, and splits nodes on a random subset of the full feature set. Predictions from multiple decision trees are averaged to obtain a single probabilistic output~\cite{Breiman2001}. The second model class, gradient boosted decision trees (GBDT), iteratively adds models to an ensemble that predict the errors of models already in the ensemble, and adds predictions from all models in the ensemble for the final prediction~\cite{Chen2016}. The third model class is the multilayer perceptron (MLP) with ReLU activations~\cite{He2015}, a feed-forward neural network in which nodes in each layer are fully-connected to the following layer. MLP models are trained with Adam optimization using an adaptive learning rate~\cite{Kingma2014}. For all models considered, hyper-parameters are tuned using a held-out validation set and models are evaluated on a held-out test set, using a 60-20-20 split into training, validation, and test sets.

\section{Results}
\label{sec:results}
\begin{figure}%
    \centering
    \subfloat[]{{\includegraphics[width=.333\textwidth, trim=.3cm .3cm .3cm .3cm]{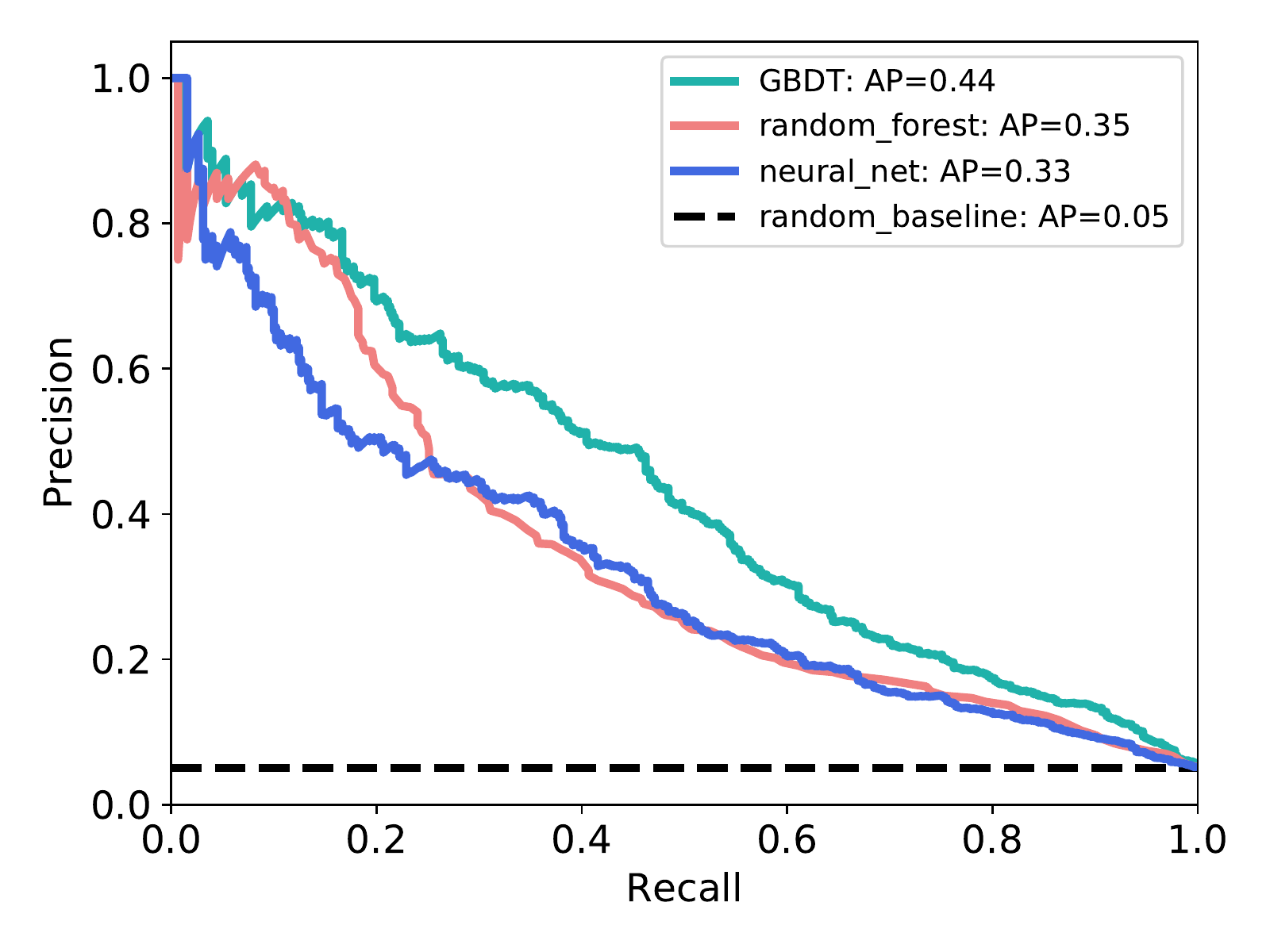} }}%
    \subfloat[]{{\includegraphics[width=.333\textwidth, trim=.3cm .3cm .3cm .3cm]{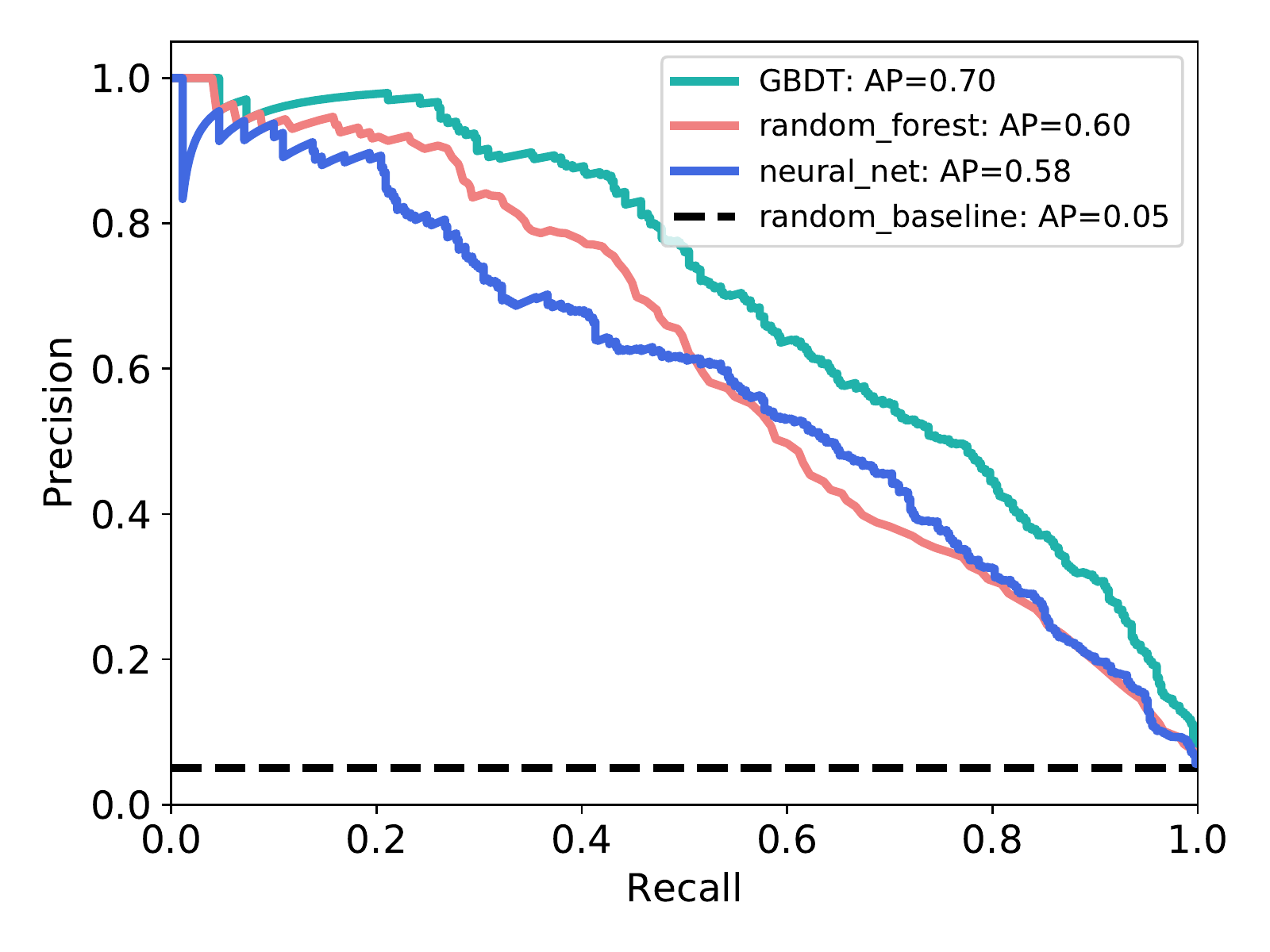} }}%
    \subfloat[]{{\includegraphics[width=.333\textwidth, trim=.3cm .3cm .3cm .3cm]{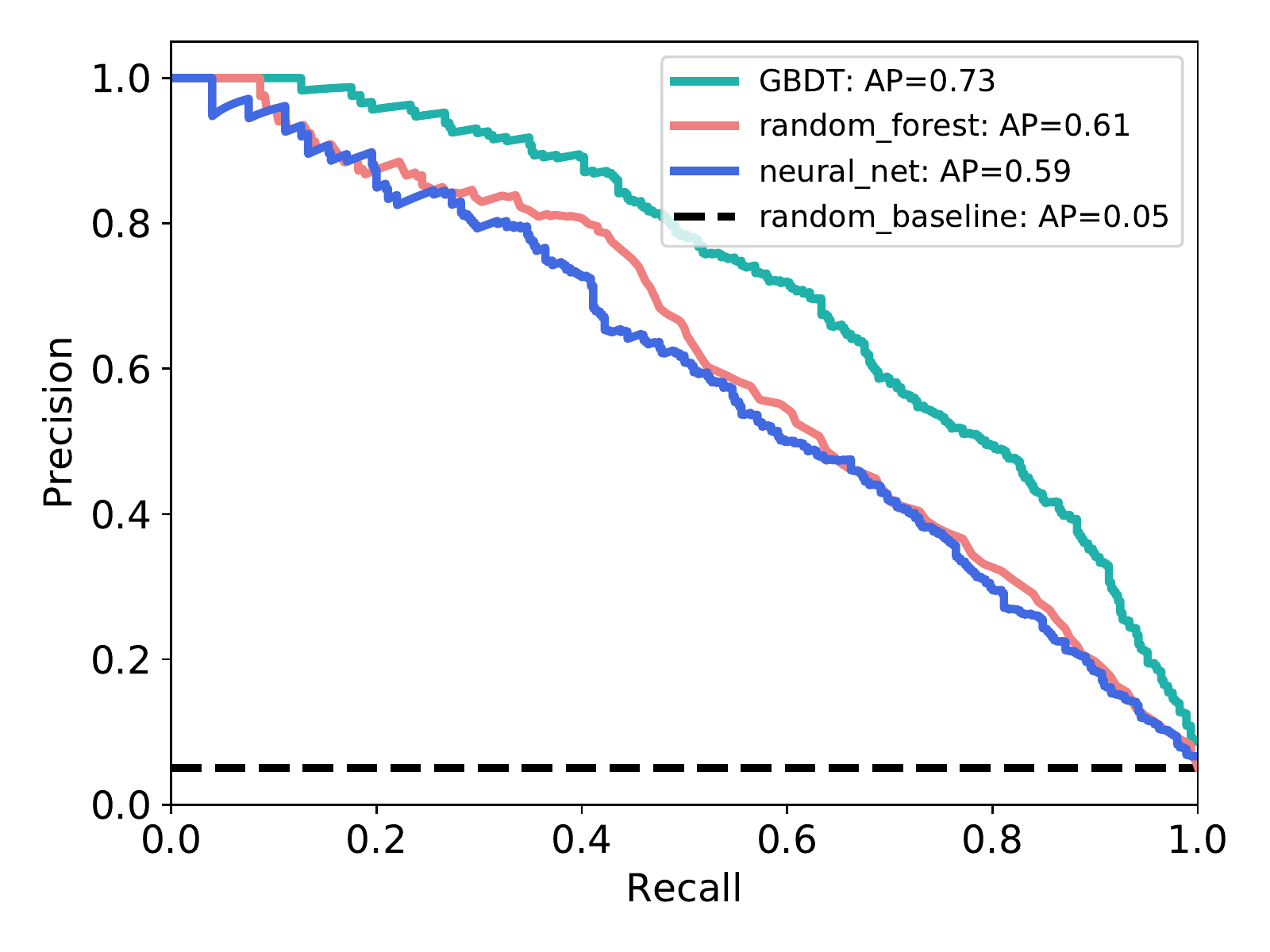} }}%
    \\\subfloat[]{{\includegraphics[width=.333\textwidth, trim=.3cm .3cm .3cm .3cm]{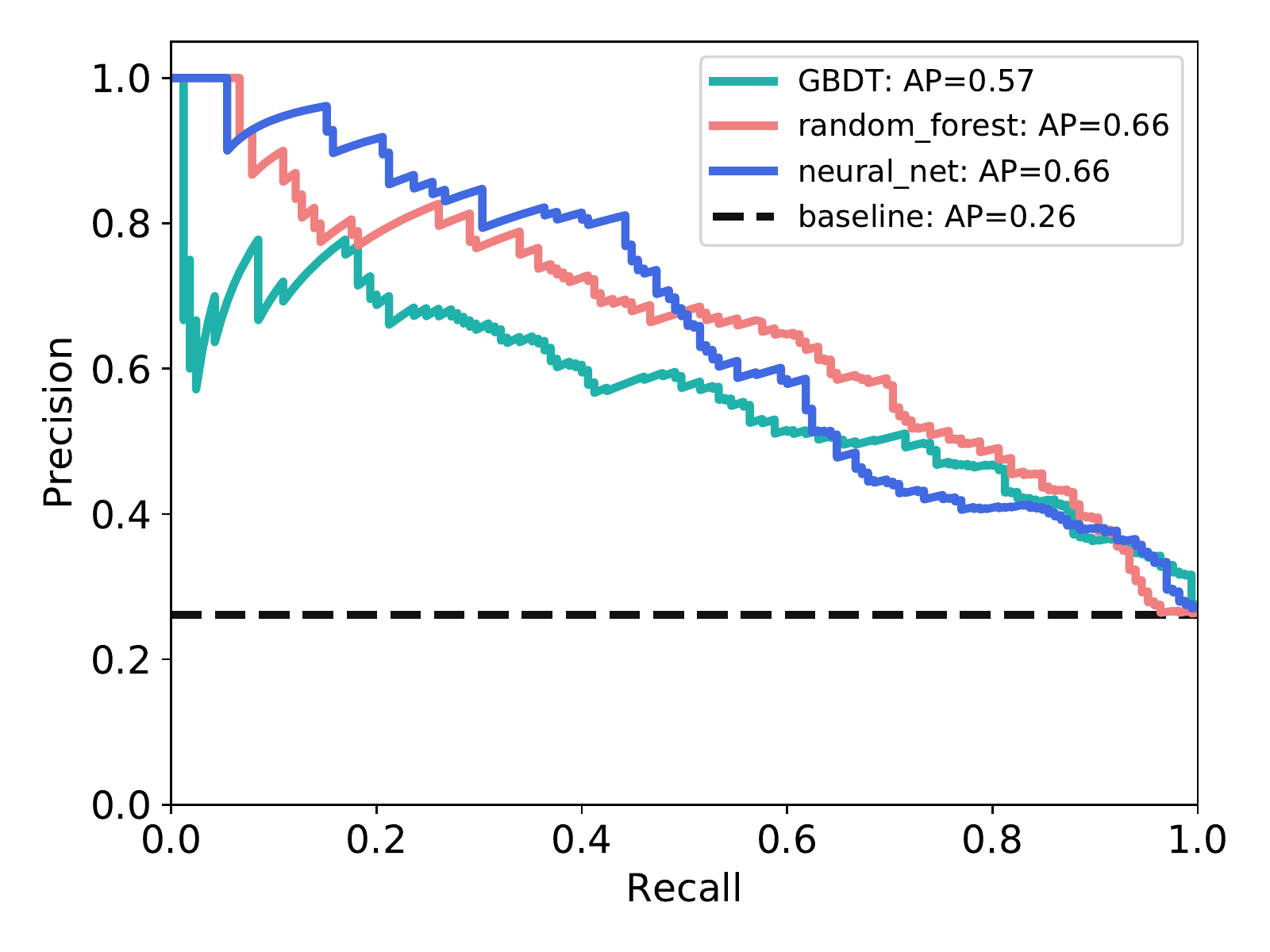} }}%
    \subfloat[]{{\includegraphics[width=.333\textwidth, trim=.3cm .3cm .3cm .3cm]{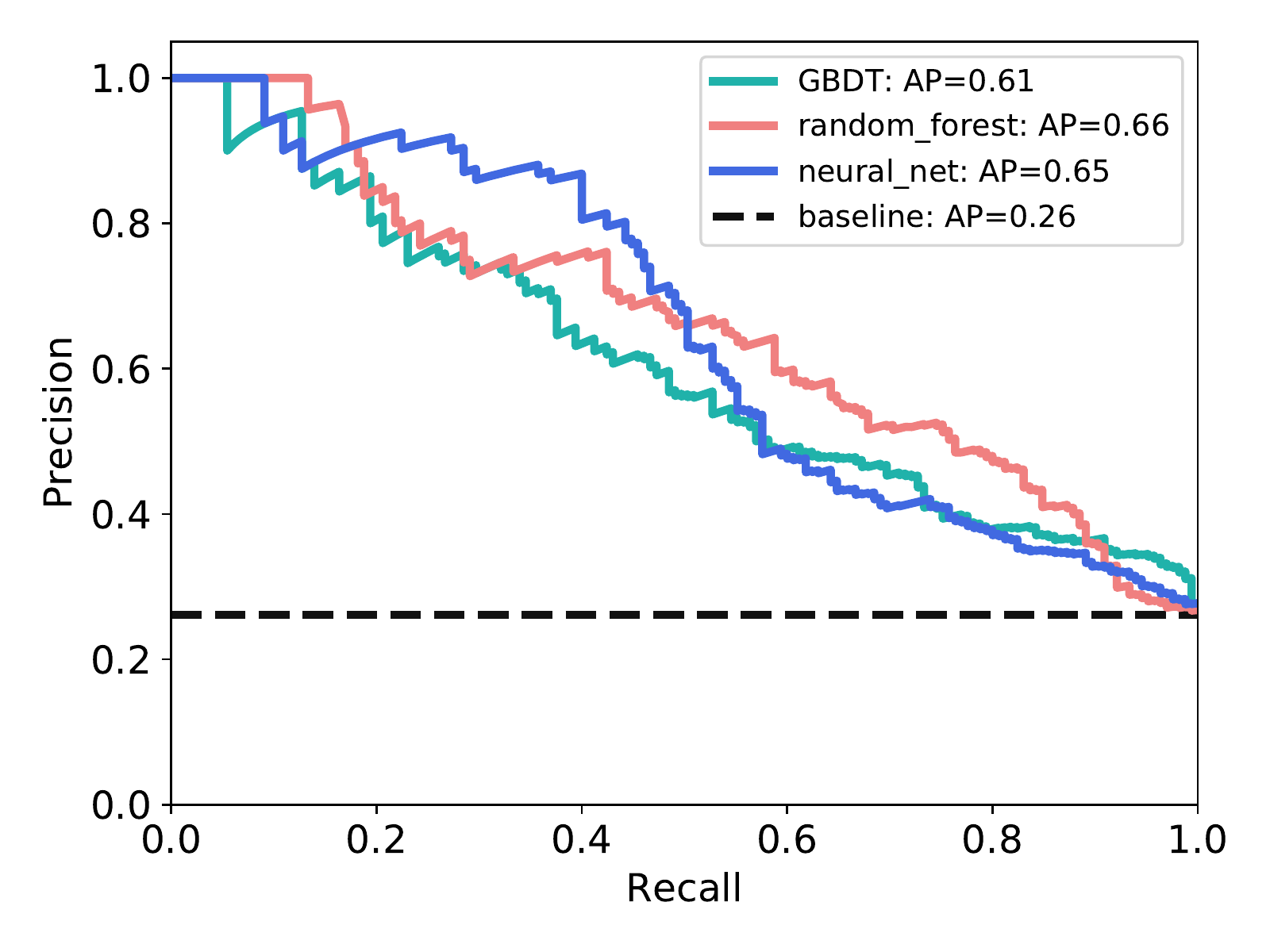} }}%
    \subfloat[]{{\includegraphics[width=.333\textwidth, trim=.3cm .3cm .3cm .3cm]{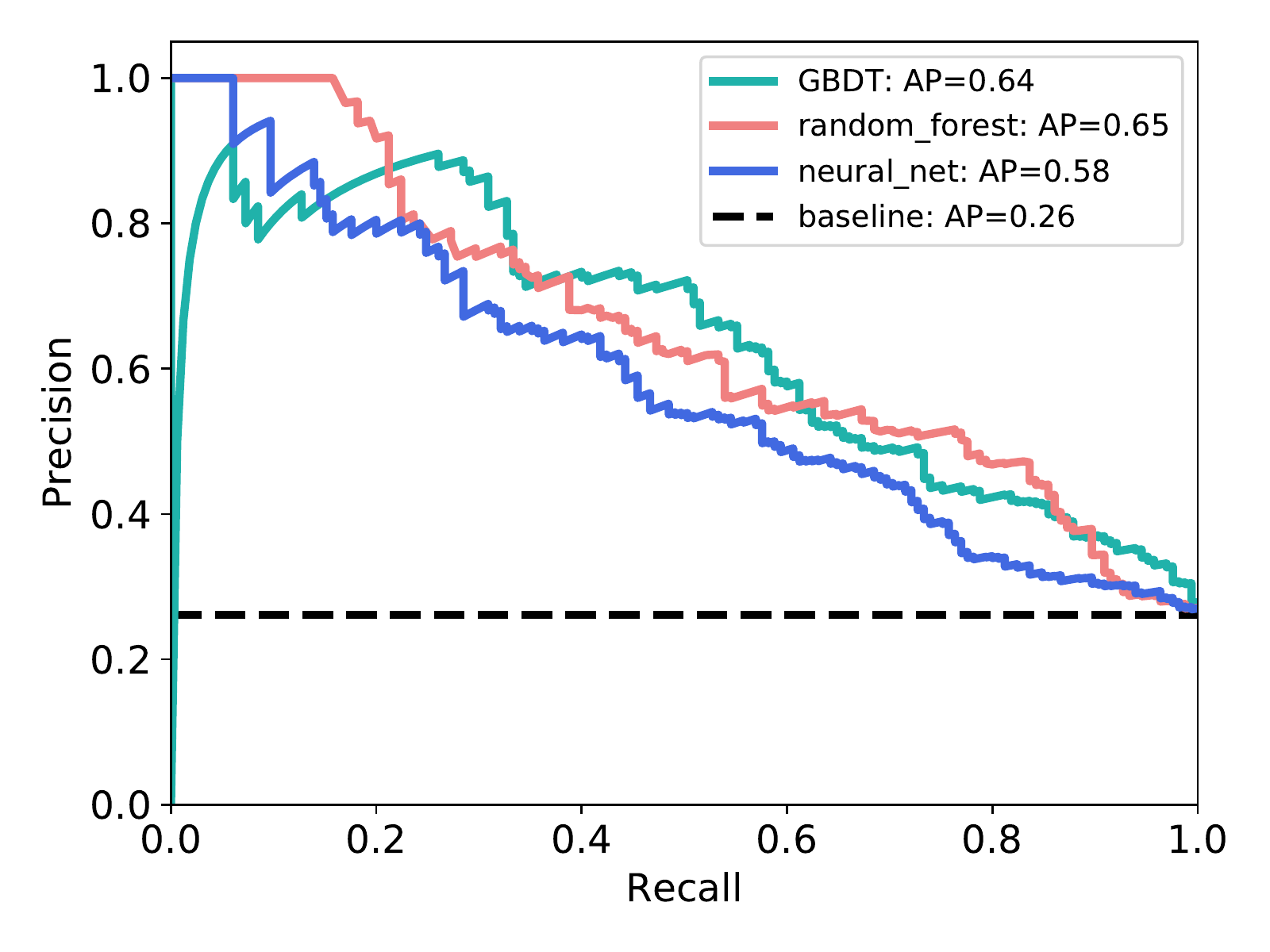} }}%
    \caption{Precision and recall curves for four model classes predicting monthly flood occurrence (top) and time to peak river level (bottom). Results are reported for three experiments: ungauged locations (left), forecasting (middle), and forecasting with a rainfall oracle (right). A gradient boosted decision tree (GBDT), random forest classifier, and multi-layer perception are compared to a random baseline. }%
    \label{fig:prec_rec}%
\end{figure}

The results of all three experiments are shown with precision-recall (PR) curves in Figure~\ref{fig:prec_rec}. PR curves allow a decision-maker to choose where to threshold the probability produced by a classifier in order to produce a desired trade-off between false positives (predicted floods that do not occur) and false negatives (actual floods that were not predicted). The machine learning models are compared to a random baseline classifier that would assign a random probability of flooding to each example, giving a PR curve that is a horizontal line at a precision corresponding to the fraction of positive examples in the test data. 

Figure~\ref{fig:prec_rec} shows that all models considered outperform the random baseline. Across all three gauge-month experiments, the GBDT is the best classifier in terms of Average Precision (AP). This is likely due to the gradient boosting on difficult-to-classify examples, which makes the GBDT particularly well-suited for an imbalanced data set of rare flooding events. For classifying time-to-peak bins, all models (excluding the random baseline) demonstrate roughly similar performance. This may be because the time-to-peak task uses a much more balanced dataset than the gauge-month experiments -- $25\%$ positive class labels compared to $~5.5\%$ positive class labels.


We also compare predictive performance to the operational flood predictions released by NOAA and collected in the Iowa Environmental Mesonet (IEM) database~\cite{IEM}. We examine predictions made between 2011 and 2019 by the Northeast River Forecast Center, which includes river gauging sites in New York, Massachusetts, Connecticut, Rhode Island, and Vermont. The NOAA flood forecasts are compared to USGS ground truth measurements and thresholded using the `minor flood' threshold, as in the gauge-month model described above. Three-day-lookahead river level forecasts at 6 hour intervals are aggregated to produce monthly flood occurrence statistics. Approximately 4,100 gauge-month predictions from 54 gauging locations are evaluated for accuracy. The results are shown in Table~\ref{tab:noaa}. While predictions from the Northeast River Forecast Center cover a different (but overlapping) region compared to our study area, this comparison suggests our models may significantly outperform NOAA's operational models on monthly flood prediction.

The finding that our models may outperform NOAA's models is unexpected because NOAA's prediction models have access to river levels as recently as three days prior to the prediction time, while our statistical models only have access to river levels from the prior month. For a precision of 0.5, our forecasting models (top row, center column of Figure~\ref{fig:prec_rec}) all achieve recall values of at least 0.6 in comparison to the 0.245 recall value that the NOAA Northeast River Forecast Center model achieves. If this trend is consistent across the contiguous U.S., our model represents a more than twofold improvement in the proportion of actual flood events that are predicted ahead of time.

\begin{table}[t]
\caption{The precision and recall values for NOAA operational predictions from the Northeast River Forecast Center.}
\label{tab:noaa}
\vskip 0.15in
\begin{center}
\begin{small}
\begin{sc}
\begin{tabular}{cc}
\toprule
Precision & Recall \\
\midrule
0.5    & 0.245\\
\bottomrule
\end{tabular}
\end{sc}
\end{small}
\end{center}
\vskip -0.1in
\end{table}

\section{Discussion \& Conclusions}
These preliminary results provide promising evidence that multi-basin flood prediction with statistical models is possible and is deserving of additional research despite the fact that such an approach has not yet been used in prior work to the best of our knowledge. We believe that the incorporation of additional remotely-sensed data streams and more sophisticated machine learning techniques has the potential to produce even higher quality multi-basin flood prediction models. Further, while this work focuses on six geographically-distributed states, adding data from stream gauge locations throughout the conterminous U.S. can potentially improve generalizability of the trained models. In summary, our work provides a proof-of-concept that multi-basin flood prediction using statistical models and remotely-sensed data has considerable value, and can overcome many of the shortcomings of physics-based flood prediction models such as reliance on time-consuming calibration to small geographic areas. Regional-scale flood susceptibility models increase the accessibility of accurate and cost-effective disaster planning for populations at risk of experiencing flooding anywhere in the world.

\section{Future Work}
Future work includes training and testing our model using data from the entire U.S., and comparing this to flood predictions from NOAA for the entire U.S.  We also plan to closely analyze the geographic areas and climatological and topographical conditions where our predictions perform significantly differently from NOAA's predictions.


We are also planning to incorporate image-based data streams and deep convolutional vision networks into our models in order to improve predictions. We hypothesize that high-resolution satellite imagery, synthetic aperture radar (SAR) remote sensing data, and information about the location of surface water could provide additional information about land cover and river topography that is not currently provided to the models and which could increase their predictive power. Additional work is also planned that uses a combination of local and global regressors inspired by \citet{gigi2019towards} and \citet{kratzert2018rainfall}. A generalizable model would still be trained on data from different basins, but would be combined with local models trained on data from the basin where the model will be applied.
 
\section*{Acknowledgements}
The authors would like to thank Guy Schumann, Theo Barnhart, Jonathan Stock, Jack Eggleston, and Baishali Chaudhury for their assistance. The authors would also like to acknowledge the NASA FDL program for supporting this research project.

\bibliographystyle{unsrtnat}
\setcitestyle{square}
\bibliography{FDL_floods}
\end{document}